\documentclass[conference]{IEEEtran}
\IEEEoverridecommandlockouts

\usepackage{cite}
\usepackage{amsmath,amssymb,amsfonts}
\usepackage{algorithmic}
\usepackage{graphicx}
\usepackage{textcomp}
\usepackage{xcolor}
\usepackage{multirow}
\usepackage{jabbrv}

\def\BibTeX{{\rm B\kern-.05em{\sc i\kern-.025em b}\kern-.08em
    T\kern-.1667em\lower.7ex\hbox{E}\kern-.125emX}}

\makeatletter 
\newcommand{\linebreakand}{%
\end{@IEEEauthorhalign}
\hfill\mbox{}\par
\mbox{}\hfill\begin{@IEEEauthorhalign}
}
\makeatother 

\begin{document}

\title{Dynamic Neural Communication: Convergence of Computer Vision and Brain--Computer Interface\\

\thanks{This work was partly supported by Institute of Information \& Communications Technology Planning \& Evaluation (IITP) grant funded by the Korea government (MSIT) (No. RS--2021--II--212068, Artificial Intelligence Innovation Hub, No. RS--2024--00336673, AI Technology for Interactive Communication of Language Impaired Individuals, and No. RS--2019--II190079, Artificial Intelligence Graduate School Program (Korea University)).}
}

\author{\IEEEauthorblockN{Ji-Ha Park}
\IEEEauthorblockA{\textit{Dept. of Artificial Intelligence} \\
\textit{Korea University} \\
Seoul, Republic of Korea \\
jiha\_park@korea.ac.kr}

\\

\IEEEauthorblockN{Soowon Kim}
\IEEEauthorblockA{\textit{Dept. of Artificial Intelligence} \\
\textit{Korea University} \\
Seoul, Republic of Korea \\
soowon\_kim@korea.ac.kr}

\and

\IEEEauthorblockN{Seo-Hyun Lee}
\IEEEauthorblockA{\textit{Dept. of Brain and Cognitive Engineering} \\
\textit{Korea University} \\
Seoul, Republic of Korea \\
seohyunlee@korea.ac.kr}

\\

\IEEEauthorblockN{Seong-Whan Lee}
\IEEEauthorblockA{\textit{Dept. of Artificial Intelligence} \\
\textit{Korea University} \\
Seoul, Republic of Korea \\
sw.lee@korea.ac.kr}
}

\maketitle

\begin{abstract}
Interpreting human neural signals to decode static speech intentions such as text or images and dynamic speech intentions such as audio or video is showing great potential as an innovative communication tool. Human communication accompanies various features, such as articulatory movements, facial expressions, and internal speech, all of which are reflected in neural signals. However, most studies only generate short or fragmented outputs, while providing informative communication by leveraging various features from neural signals remains challenging. In this study, we introduce a dynamic neural communication method that leverages current computer vision and brain-computer interface technologies. Our approach captures the user's intentions from neural signals and decodes visemes in short time steps to produce dynamic visual outputs. The results demonstrate the potential to rapidly capture and reconstruct lip movements during natural speech attempts from human neural signals, enabling dynamic neural communication through the convergence of computer vision and brain--computer interface.
\end{abstract}


\section{INTRODUCTION}
Brain--computer interface (BCI) is a technology that interprets brain signals, capturing various aspects of the user's intentions and mental states~\cite{jeong2019classification, chaudhary2016brain, han2020classification, karikari2023review, lee2020continuous}. There has been growing interest in using this technology to assist communication by decoding and delivering users' thoughts in diverse forms~\cite{anumanchipalli2019speech}. Brain--to--speech (BTS) technology, which directly conveys a user's intentions, has emerged as a significant new form of neural communication~\cite{lee2020neural}. The BTS system aims to decode both external and internal speech--related neural signals and generate speech outputs in formats such as text or audio. Speech BCIs such as BTS not only serve as assistive and rehabilitative neuroprosthetics for individuals with speech impairments but also demonstrate significant potential for broader development as a novel neural communication system for general users~\cite{moses2021neuroprosthesis, lee2022toward, willett2023high, prabhakar2020framework}.

In parallel, advancements in computer vision (CV) are enhancing the generation of talking faces through lip-sync methods and decoding speech intentions through lip reading using text, images, and audio~\cite{zhou2018visemenet, kim2018discriminative, prajwal2020lip, lee2020uncertainty, ji2021audio}. When combined with BCIs, these CV technologies can offer users a more immersive and realistic visual communication experience~\cite{bai2023dreamdiffusion, takagi2023high}. Consequently, there is increasing interest in decoding facial expressions and articulatory movements from neural signals to generate dynamic visual outputs~\cite{chartier2018encoding}. These are beginning to explore the representation of visual speech intentions from neural signals, an area that has not been extensively studied yet. Further efforts are required to generate multimodal--based dynamic and realistic outputs, such as talking faces or avatars~\cite{metzger2023high}.

However, despite these promising advancements, most multimodal BCI research based on neural signals still faces challenges in visually reconstructing and dynamically presenting users' speech intentions. Unlike more conventional and intuitive data such as text, speech, and images, capturing key features of complex facial expressions and subtle lip movements from noisy and limited neural signals remains a significant challenge. Decoding rapid and subtle movements in tasks that involve reconstructing continuous facial articulations remains challenging. While invasive methods offer the potential for improved reconstruction quality due to their high spatial resolution, they involve surgical risks and limited practicality~\cite{willett2023high, metzger2023high}. Non--invasive methods allow for faster reconstruction due to their high temporal resolution, but high--quality reconstructions remain elusive due to the low signal--to--noise ratio~\cite{lee2023AAAI}. Therefore, in--depth research is required to develop approaches that can reconstruct high--quality, realistic visual outputs using non--invasive methods.

\begin{figure*}[t]
\centerline{\includegraphics[width=0.99\textwidth]{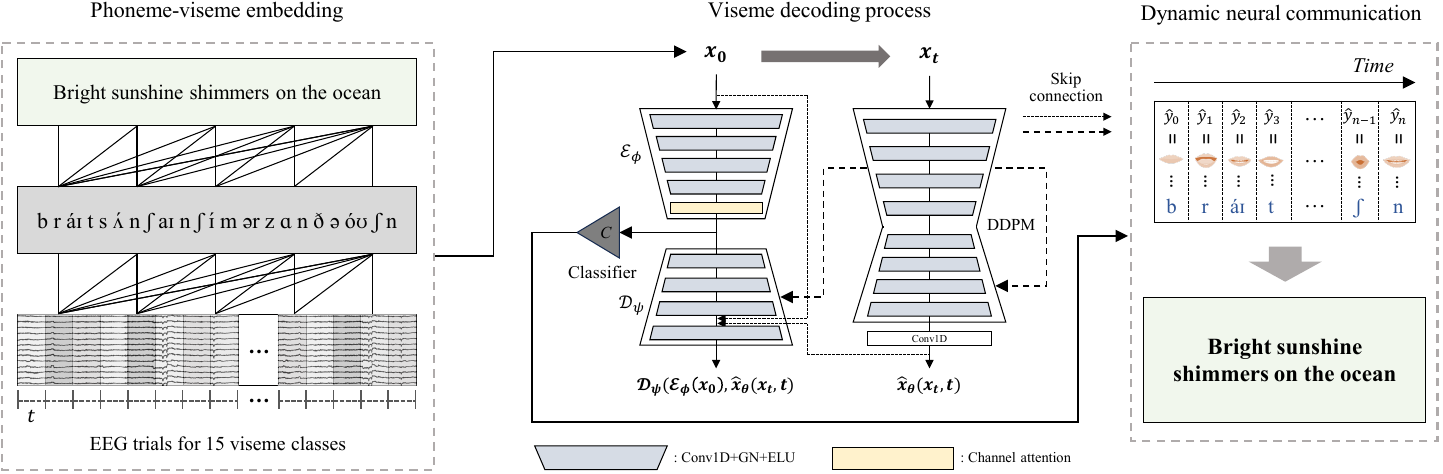}}
\caption{Overall architecture of the sentence--based viseme decoding framework from overt speech EEG. The EEG signals from spoken sentences are segmented based on phoneme intervals, and the units are mapped to condensed viseme classes. The embedded features are adjusted to variable time lengths \textit{t} and used for training. The diffusion--based EEG signal decoding model is trained to effectively capture the viseme information. Finally, the predicted viseme sequences are sequentially reconstructed to form a complete sentence.}
\label{fig1}
\end{figure*}

In this paper, we introduce a decoding framework from electroencephalography (EEG) and electromyography (EMG) signals for dynamic neural communication. This framework seeks to capture rapidly changing lip movements based on speech--related neural signals and reconstruct them like natural speech. By redefining phonemes, the smallest units of speech, as condensed viseme classes, we have improved decoding efficiency~\cite{cappelletta2012phoneme}. This approach enables fast and realistic neural communication using non--invasive neural signals and contributes to the generation of dynamic visual outputs, offering a novel means of face--to--face interaction.

\section{METHODS}

\subsection{Dataset Description}
An EEG cap was used to measure EEG signals from a total of 128 channels, including the reference electrode, and an additional 10 channels of EMG signals were measured using separate electrodes. The EMG channels were evenly distributed to capture facial movements and muscle signals around the mouth. For in--depth analysis, overt speech data from a single subject, a healthy male native English speaker, were used, as this subject provided distinct signal recordings. A total of 474 sentences were selected from the MOCHA--TIMIT corpus for their distinct characteristics and recorded. These were initially divided into 424 sentences for the training set and 50 sentences for the test set. The study was conducted by the Declaration of Helsinki and approved by the Korea University Institutional Review Board [KUIRB-2022-0079-03]. The EEG signals were recorded using Brain Vision/Recorder (Brain--Product GmbH, Germany), and the corresponding audio signals were also recorded.

For all sentences, phonemes were segmented based on the real spoken audio to align the viseme classes assigned to each time interval~\cite{mcauliffe2017montreal}. Each phoneme unit was condensed into one of 15 viseme classes, allowing for the reconstruction of lip movements with simplified classes~\cite{lim2000text, pandzic2003mpeg}. The segmented units with varying time lengths were adjusted to a uniform length for the training process and divided into short segments of 64~ms, 128~ms, and 256~ms. The training and test sets consisted of a total of 13,826 and 1,646 trials, respectively.

\subsection{Signal Preprocessing}
We applied a 5th--order Butterworth bandpass filter within the range of 30--499 Hz, which was chosen to encompass the wide frequency range associated with speech and articulatory information~\cite{lee2020neural}. Additionally, notch filtering was performed to remove line noise at harmonics of 60~Hz. Signal preprocessing was primarily conducted in Matlab, with Python used as needed. The main preprocessing tool utilized was the EEGLAB Toolbox~\cite{delorme2004eeglab}.

\subsection{Viseme Decoding Model}
The main backbone architecture and training details followed Kim et al.~\cite{kim2023diff}. Unlike the conventional diffusion--based model approach, which predicts noise at each step, this model had been specifically adapted for EEG data decoding by progressively adding noise to reconstruct the original EEG signal, as shown in Fig.~\ref{fig1}. To achieve this, denoising diffusion probabilistic models (DDPM) based on a time--conditional U--Net architecture were used~\cite{ho2020denoising}. Additionally, a conditional autoencoder composed of an encoder $\mathcal{E}\phi$ and a decoder $\mathcal{D}\psi$ was employed to compensate for the information loss of the DDPM~\cite{zhang2022unsupervised}. To indirectly incorporate the effect of the DDPM, $\mathcal{D}\psi$, instead of the output of $\mathcal{E}\phi$, was skip--connected to the DDPM layers, and both the original signal and the DDPM output were also skip--connected to the stage immediately preceding the final layer of $\mathcal{D}\psi$. To effectively decode the characteristics of lip movements, channel attention was applied after the final layer block of $\mathcal{E}\phi$. The channel attention layer learned the importance of each channel and multiplied the original input by the differently assigned weights to produce an improved output. $\mathcal{E}\phi$ extracted meaningful features related to visemes, and the latent vector, represented as a one--dimension through an adaptive average pooling layer, was passed through the classifier $\mathcal{C}\rho$ to predict the final class. $\mathcal{C}_\rho$ was fine--tuned based on the Kolmogorov–Arnold Networks to optimize the activation functions, contributing to improved prediction performance~\cite{liu2024kan}.

\begin{table}[t]
\setlength{\tabcolsep}{2pt}
\renewcommand{\arraystretch}{1.25}
\caption{Top--1 and top--3 accuracy, F1--score and AUC score of viseme decoding for overt speech neural signals.}
\begin{center}
\begin{tabular}{cccccc}
\hline
\textbf{}                         &        & \textbf{\begin{tabular}[c]{@{}c@{}}Top--1\\ Acc. (\%) $\uparrow$\end{tabular}} & \textbf{\begin{tabular}[c]{@{}c@{}}Top--3\\ Acc. (\%) $\uparrow$\end{tabular}} & \textbf{F1--score $\uparrow$} & \textbf{AUC (\%) $\uparrow$} \\ \hline
\multirow{3}{*}{\textbf{EEG+EMG}} & 64 ms  & 31.81                                                              & 56.44                                                              & 27.73             & 78.38             \\
                                  & 128 ms & 33.77                                                              & 56.54                                                              & 31.17             & 80.98             \\
                                  & 256 ms & 33.55                                                              & 57.59                                                              & 32.09             & 81.53             \\ \hline
\multirow{3}{*}{\textbf{EEG}}     & 64 ms  & 27.33                                                              & 54.09                                                              & 22.96             & 75.03             \\
                                  & 128 ms & 27.73                                                              & 52.35                                                              & 24.34             & 77.14             \\
                                  & 256 ms & 29.39                                                              & 52.82                                                              & 25.30             & 77.70             \\ \hline
\end{tabular}
\label{tab1}
\end{center}
\footnotesize{$^*$Acc.: accuracy}
\end{table}

\subsection{Viseme--to--sentence Reconstruction}
The phoneme--segmented and condensed viseme sequences underwent an additional reconstruction process to align with the original sentence. We conducted tests on a dataset of 50 sentences, where the pre--trained model was designed and applied to guide the discrete results into a coherent sentence. To effectively capture long--term dependencies in the sequence data, we employed a long short--term memory (LSTM) model, which has demonstrated strong performance in time--series analysis and natural language processing tasks. This was trained with ground truth viseme sequences to ensure that the predicted viseme sequences could accurately guide the reconstruction of the target sentences.

\section{RESULTS AND DISCUSSION}

\subsection{Viseme Decoding Performance}
Table~\ref{tab1} displays the quantitative performance of viseme decoding from overt speech. It compares the results when using only EEG signals and when using both EEG and EMG signals. The neural signals, epoched from actual speech audio, were analyzed for three other time lengths: 64~ms, 128~ms, and 256~ms. The classification results for 15 viseme classes were quantified, and the evaluation metrics used were accuracy, F1--score, and area under the curve (AUC). To gain a broader understanding of the classification performance, both top--1 and top--3 accuracy were recorded. When both EEG and EMG signals were combined, for the 64~ms time length, the top--1 accuracy, top--3 accuracy, F1--score, and AUC score were 27.33, 54.09, 22.96, and 75.03, respectively. For the 128~ms time length, the top--1 accuracy, top--3 accuracy, F1--score, and AUC score were 27.73, 52.35, 24.34, and 77.14, respectively. For the 256~ms time length, the top--1 accuracy, top--3 accuracy, F1--score, and AUC score were 29.39, 52.82, 25.30, and 77.70, respectively. By incorporating EMG signals alongside EEG, we enhanced the decoding performance, compensating for the limitations of noisy EEG signals. Muscle signals, such as facial and lip movements, strongly contributed to improving decoding performance, proving that articulatory movement information has a significant impact on viseme decoding.

Additionally, when using only EEG signals, for the 64~ms time length, the top--1 accuracy, top--3 accuracy, F1--score, and AUC score were 27.33, 54.09, 22.96, and 75.03, respectively. For the 128~ms time length, the top--1 accuracy, top--3 accuracy, F1--score, and AUC score were 27.73, 52.35, 24.34, and 77.14, respectively. For the 256~ms time length, the top--1 accuracy, top--3 accuracy, F1--score, and AUC score were 29.39, 52.82, 25.30, and 77.70, respectively. Although performance slightly declined when using EEG signals alone without EMG signals, the difference was not significant, demonstrating that it is still possible to decode lip shape information using only brain signals. These results suggest the potential to decode visemes not only from overt EEG but also from mimed or imagined EEG, showing promise for applications such as neuroprosthesis for speech impairments and the development of new forms of neural communication~\cite{willett2023high}.

\begin{figure}[t]
\centerline{\includegraphics[width=0.99\columnwidth]{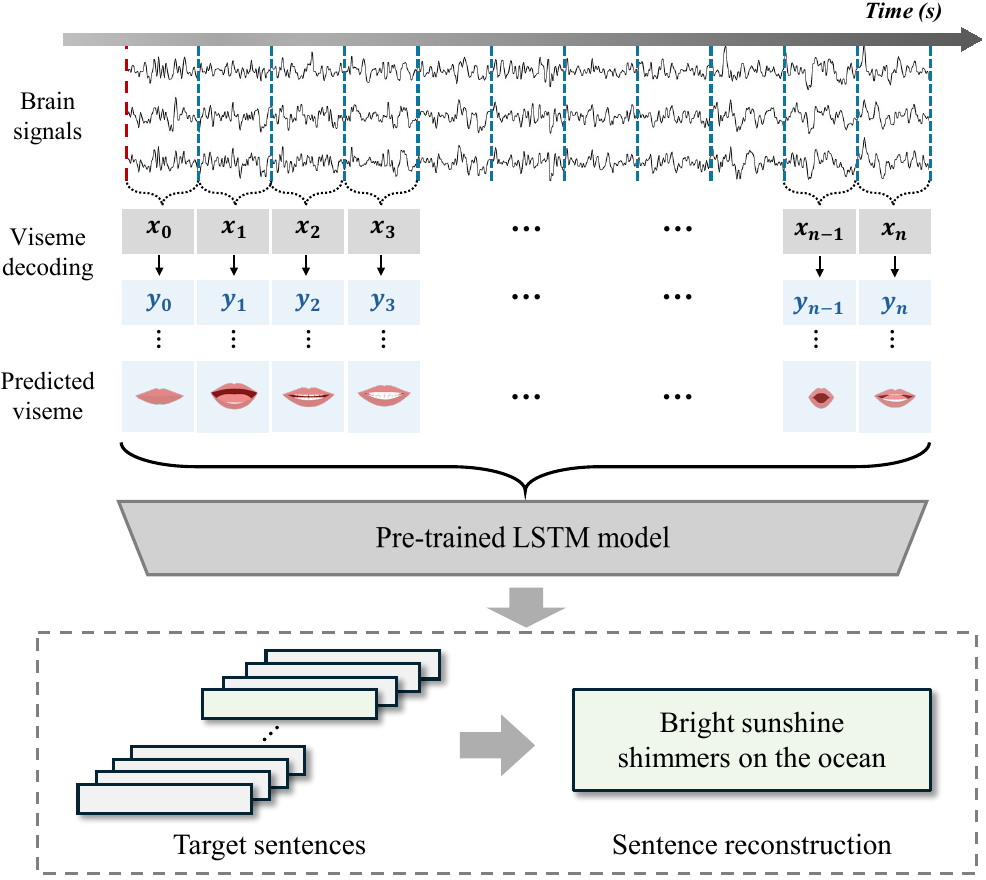}}
\caption{The predicted labels from EEG segments are arranged into a one--dimensional viseme sequence. The incomplete sequence is reconstructed into a complete sentence using a pre--trained LSTM model. The LSTM model was trained using the original viseme sequences for 50 predefined sentences.}
\label{fig2}
\end{figure}

\subsection{Visual Interaction from Neural Signals}
We analyzed the potential to capture and reconstruct lip movements within short time intervals during natural sentence--level speech. As shown in Table~\ref{tab1}, we compared performance by mapping phonemes to visemes and adjusting the time length of the corresponding segments in sentence data. While using only EEG signals, the top--1 accuracy was highest for segments with the longest time length of 256~ms, but overall, shorter time lengths showed similar performance levels without significant degradation. Although decoding performance slightly decreased for concise data under 64~ms due to the limited information available, we observed fairly high viseme decoding results for moderately short time lengths. This demonstrated the ability to capture rapidly changing lip movements and proved the potential for efficiently reconstructing visual speech intentions, serving as a substitute for more granular decoding approaches that use a larger number of classes, such as phonemes.

As shown in Fig.~\ref{fig2}, the discretely derived viseme results were represented as a one--dimensional array. Inference for 50 sentences in the test set was conducted using a pre--trained LSTM model to match the target sentence. As a result, the variety of visemes decoded in short time intervals could be reconstructed into the original sentences, and all were accurately inferred within the pre--defined sentence set. This demonstrated the potential to overcome the limitations of decoding precise and detailed speech intentions from neural signals. Ultimately, expanding through the convergence of CV technologies could pave the way toward more universal and dynamic neural communication.

\section{CONCLUSION}
We proposed viseme decoding framework that enables visual neural communication. In speech attempts involving long sentences, similar to those in real conversations, phonemes were segmented and labelled with condensed visemes to enhance decoding efficiency. Each segment was trained using short time intervals within the proposed diffusion--based model, and the feasibility of the decoding process was demonstrated. Our viseme decoding--based method enabled the reconstruction of pre--defined sentences. Moreover, by converging CV technologies with BCI, this approach demonstrated the potential to reconstruct dynamic visual outputs based on lip shape decoding, enabling lively and engaging communication. The results of this study offer promising potential to advance neural communication beyond fragmented and simple forms, moving towards more continuous, realistic, and intuitive interactions. To further build on these advancements, future works could focus on reconstructing decoded visemes into continuous images, synchronized with real voice, and lip--synced to generate realistic talking faces or avatars.

\bibliographystyle{jabbrv_IEEEtran}
\bibliography{REFERENCE_jh}

\end{document}